\documentclass{template}

\usepackage{microtype}
\usepackage{graphicx}
\usepackage{subfigure}
\usepackage{booktabs} 

\usepackage{hyperref}

\usepackage{amsmath}
\usepackage{amssymb}
\usepackage{mathtools}
\usepackage{amsthm}
\usepackage{times}

\usepackage{fancyhdr}
\usepackage{xcolor} 
\usepackage{algorithm}
\usepackage{algorithmic}
\usepackage{eso-pic} 
\usepackage{forloop}
\usepackage{url}
\usepackage[capitalize,noabbrev]{cleveref}

\usepackage{environ}
\newcommand{\acksection}{\section*{Acknowledgments and Disclosure of Funding}}
\NewEnviron{ack}{%
  \acksection
  \BODY
}

\theoremstyle{plain}

\theoremstyle{definition}

\theoremstyle{remark}

\usepackage[textsize=tiny]{todonotes}

\makeatletter
\DeclareRobustCommand{\cev}[1]{%
  {\mathpalette\do@cev{#1}}%
}
\newcommand{\do@cev}[2]{%
  \vbox{\offinterlineskip
    \sbox\z@{$\m@th#1 x$}%
    \ialign{##\cr
      \hidewidth\reflectbox{$\m@th#1\vec{}\mkern4mu$}\hidewidth\cr
      \noalign{\kern-\ht\z@}
      $\m@th#1#2$\cr
    }%
  }%
}
\makeatother

\newcommand{\pg}{\mathbf{p}}
\newcommand{\qg}{\mathbf{q}}
\newcommand{\Sg}{\mathbf{S}^{\text{g}}}

\newcommand{\vm}{\mathbf{v}}
\newcommand{\va}{\mathbf{\theta}}
\newcommand{\V}{\mathbf{V}}

\newcommand{\pf}{{\mathbf{f}}}
\newcommand{\qf}{{\mathbf{f}}^{\text{q}}}
\newcommand{\Sf}{{\mathbf{S}}^{\text{f}}}

\newcommand{\wmsoc}{\mathbf{w}}
\newcommand{\wrsoc}{\mathbf{w}^{\text{r}}}
\newcommand{\wisoc}{\mathbf{w}^{\text{i}}}

\newcommand{\pd}{\text{p}^{\text{d}}}
\newcommand{\qd}{\text{q}^{\text{d}}}
\newcommand{\Sd}{\text{S}^{\text{d}}}
\newcommand{\pgmin}{\underline{\text{p}}}
\newcommand{\pgmax}{\overline{\text{p}}}
\newcommand{\qgmin}{\underline{\text{q}}}
\newcommand{\qgmax}{\overline{\text{q}}}
\newcommand{\vmmin}{\underline{\text{v}}}
\newcommand{\vmmax}{\overline{\text{v}}}

\begin{document}

\title{Efficiently Training Deep-Learning Parametric policies using Lagrangian Duality}

\author{\IEEEauthorblockN{Andrew W. Rosemberg\IEEEauthorrefmark{1},
Alexandre Street\IEEEauthorrefmark{2},
Davi M. Vallad\~ao\IEEEauthorrefmark{2} and
Pascal Van Hentenryck\IEEEauthorrefmark{1}}
\IEEEauthorblockA{\IEEEauthorrefmark{1}\textit{Georgia Institute of Technology}, Atlanta, GA, United States\\
\{arosemberg, vanhentenryck\}@gatech.edu}
\IEEEauthorblockA{\IEEEauthorrefmark{2}\textit{Pontifical Catholic University of Rio de Janeiro}, Rio de Janeiro, RJ, Brazil\\
\{alexandre.street, davi.valladao\}@puc-rio.br}
}








\maketitle

\begin{abstract}
Constrained Markov Decision Processes (CMDPs) are critical in many
high-stakes applications, where decisions must optimize cumulative
rewards while strictly adhering to complex nonlinear constraints. In
domains such as power systems, finance, supply chains, and precision
robotics, violating these constraints can result in significant
financial or societal costs. Existing Reinforcement Learning (RL)
methods often struggle with sample efficiency and effectiveness in
finding feasible policies for highly and strictly constrained CMDPs,
limiting their applicability in these environments. Stochastic dual
dynamic programming is often used in practice on convex relaxations of
the original problem, but they also encounter computational challenges
and loss of optimality. This paper introduces a novel approach,
Two-Stage Deep Decision Rules (TS-DDR), to efficiently train
parametric actor policies using Lagrangian Duality. TS-DDR is a
self-supervised learning algorithm that trains general decision rules
(parametric policies) using stochastic gradient descent (SGD); its
forward passes solve {\em deterministic} optimization problems to find
feasible policies, and its backward passes leverage duality theory to
train the parametric policy with closed-form gradients. TS-DDR
inherits the flexibility and computational performance of deep
learning methodologies to solve CMDP problems. Applied to the
Long-Term Hydrothermal Dispatch (LTHD) problem using actual power
system data from Bolivia, TS-DDR is shown to enhance solution quality
and to reduce computation times by several orders of magnitude when
compared to current state-of-the-art methods.
\end{abstract}


\begin{ack}
This research is partly funded by NSF award 2112533.

The work of Alexandre Street and Davi Valladão was partially supported by CAPES, CNPq and FAPERJ.
\end{ack}

\section{Introduction}

Sequential Decision Making under Uncertainty (SDMU) is fundamental for
realistic applications as they model real-world scenarios where
decisions unfold sequentially, shaped by an uncertain process and
previous choices \cite{shapiro2009lectures,
  powell2016unified}. Decisions at a particular time step (also known
as a stage) must account for information just revealed as well as
future uncertainty, reflecting the dynamic nature of real-world
scenarios. Ideally, this solution process returns a policy mapping
available information about the environment to
decisions. The application of SDMU extends to various domains,
including (stochastic predictive) control \cite{mesbah2016stochastic},
finance \cite{valladao2019time}, supply chains
\cite{powell2003stochastic}, and power systems
\cite{maceiral2018twenty, psr_software_2019}. 

For applications modeled as Markov Decision Processes (MDPs),
Reinforcement Learning (RL) strategies search for a policy that
maximizes the expected cumulative reward
\cite{szepesvari2022algorithms}.  Model-free RL \cite{DQN,TRPO}, which
learns policies directly from data interaction, has demonstrated
remarkable success in areas such as playing games \cite{le2022deep,
  silver2018general} and Language Models \cite{kaufmann2023survey}.

In many domains, SDMU applications feature hard nonlinear constraints
(e.g., power balance in energy systems). Constrained Reinforcement
Learning (CRL) methods search for policies satisfying the constraints
and maximizing the cumulative reward. Current CRL approaches fall into
two primary categories: primal-dual methods and feasible region
methods. Primal-dual methods convert constrained optimization into an
unconstrained dual problem by introducing Lagrangian multipliers that
are used to penalize constraint violations \cite{paternain2022safe,
  stooke2020responsive, zhang2020first,
  altman2021constrained}. Feasible region methods ensure that policy
updates stay within a feasible region, which is often determined by
trust regions \cite{achiam2017constrained, yang2020projection,
  fujimoto2019benchmarking, xuan2023constrained}.  Despite great
progress \cite{liu2022constrained}, existing approaches still struggle
in high-dimentional applications with hard nonlinear constraints,
where sampling efficiency is of paramount importance
\cite{dulac2021challenges}.

Model-Based Reinforcement Learning (MBRL), which incorporates learned
or known models of transition dynamics, can significantly improve
sample efficiency by simulating state transitions and planning within
the model \cite{deisenroth2011pilco, luo2018algorithmic, chua2018deep,
  kurutach2018model, heess2015learning, wang2019benchmarking,
  sikchi2022learning}. This aligns well with the needs of many MDP
applications where approximate models of the environment are available
\cite{thomas2021safe}. In constrained environments, MBRL techniques
have been adapted to account for safety requirements. Liu et
al. \cite{liu2020safe} present an MBRL framework that learns both
system dynamics and cost functions, optimizing policy updates based on
rollouts using the learned model. However, their approach requires
significant data collection with random policies, which can be risky
in practical applications. Cowen et al. \cite{cowen2022samba} employ
PILCO \cite{deisenroth2011pilco} alongside a Conditional
Value-at-Risk (CVaR) constraint, emphasizing risk-averse policies,
while Thomas et al. \cite{thomas2021safe} introduce penalized reward
functions combined with Soft Actor-Critic (SAC) for safe RL. While
these approaches improve constrained policy learning, they often
struggle with high computational costs and challenges in handling
complex, high-dimensional constraints effectively
\cite{jayant2022model}. This limitation is especially pronounced in
sensitive applications (e.g., power grids, financial trading,
safety-aware robotics), where constraint violations have high
penalties.



When the problem is convex and under the stage-wise independence
assumption, polices for CMDPs can be efficiently trained using dynamic
programming methods, e.g., the state-of-the-art Stochastic Dual
Dynamic Programming (SDDP) algorithm \cite{pereira1991multi}. SDDP and
similar model-based algorithms have been successfully used in many
large-scale applications, especially in energy systems
\cite{psr_software_2019}, where competing approaches have failed. In
particular, SDDP leverages duality theory to optimize the cumulative
objective while respecting constraints with the minimum number of
environment evaluations. Although SDDP and similar stochastic
programming approaches guarantee optimality and solution efficiency,
they have some inherent limitations: the stage-wise independence
assumption does not hold in general and, more importantly, they do not
return a policy that can be run in real-time.  To address this
limitation, Two-Stage Linear Decision Rules (TS-LDRs) have been
proposed. TS-LDRs remove the need for the stage-wise independence
assumption \cite{bodur2022two, nazare2023solving} and compute a policy
that is a linear function of past observations and states. TS-LDRs are
easily trained and have fast inferences, which is ideal for
time-restricted applications.  However, linear policies are rarely
suitable for the non-convex environments. Attempts to address the
non-convexity nature of the underlying problems through convex
approximations and relaxations, as necessitated by SDDP or TS-LDRs,
frequently result in oversimplifications and hence sub-optimal
decisions in real-world applications \cite{rosemberg2021assessing}.

This paper aims at bridging the gap between RL policies and stochastic
programming. It proposes the concept of {\em Two-Stage General
  Decision Rules} (TS-GDR) that generalizes TS-LDR to arbitrary
policies. Moreover, the paper explores one of its instantiations
(TS-DDR) where the policy is represented by a deep neural
network. TS-DDR is trained in a self-supervised manner using
stochastic gradient descent (SGD). In particular, during training, its
forward pass solves a multi-step {\it deterministic} optimization
problem ensuring feasibility and near-optimality, while its backward
pass leverages duality theory to obtain closed-form gradients for
updating the policy parameters. TS-DDR can be seen as a feasible
region method as it projects policy decisions back into the feasible
region. It can also be seen as a MBRL primal-dual method, since it
penalizes the distance from the feasible region in the policy
update. In that sense, TS-DDR uses a similar framework as in
\cite{agrawal2019differentiable}, but {\em one of its key benefits is
  that it does not necessitate expensive implicit function
  differentiations} needed in that approach.



To demonstrate the benefits of TS-DDR, the paper considers the {\it
  Long-Term Hydrothermal Dispatch (LTHD) problem}
\cite{maceiral2018twenty}, a CMDP where energy storage and dispatch
must be optimized over multiple time stages to minimize costs while
meeting operational constraints. Constraint satisfaction is a prime
consideration, since constraint violations, e.g., not balancing demand
and production, has extreme monetary (e.g. ruined transformers,
transmission lines) and societal (e.g. loss of power at a hospital)
costs. In addition to engineering constraints, the LTHD problem
includes challenging non-convex constraints capturing the physical
laws of AC power flow. The forms of these constraints make it
difficult for traditional methods such as SDDP to find optimal
policies. This paper applies TS-DDR to approximated models (e.g.,
DC-OPF) and exact models (e.g., AC-OPF) of the environment.

The effectiveness of TS-DDR is demonstrated on realistic instances of
the Bolivian energy system. The computational results show that TS-DDR
significantly improves solution quality compared to TS-LDR with
similar training times. They also show that TS-DDR, on large LTHD
instances, provides orders of magnitude improvements in training and
inference times, as well as solutions of higher quality compared to
SDDP on actual non-convex instances. The paper also demonstrates that
model-free RL struggles to find feasible solutions under high-penalty
constraints. 

The main contributions of this paper can thus be summarized as
follows. (1) The paper proposes TS-DDR, a novel self-supervised
learning algorithm for multi-stage stochastic optimization
problems that returns a non-linear, time-invariant, policy that can be
run in real time. TS-DDR is an instantiation of TS-GDR, a general
scheme for decision rules. (2) TS-DDR is trained both in style of
TS-LDR and RL. Using training ideas from TS-LDR, TS-DDR captures the
hard constraints of the application at hand through traditional
stochastic optimization techniques. Using training ideas from RL and
recurrent neural networks, TS-DDR returns time-invariant policies that
generalize beyond the finite horizon used during training and are
extremely fast to evaluate, addressing a fundamental limitation of
many multi-stage optimization methods. (3) TS-DDR is trained using
SGD, leveraging duality theory to obtain close-form gradients. This
avoids the need for implicit differentiation techniques. (4) TS-DDR
has been applied to realistic instances of the LTHD for the Bolivian
energy system. The results show that TS-DDR improves the
state-of-the-art on large-scale realistic instances, both in terms of
solution quality as well as training and inference times.

\section{Background} \label{sec: background}

This paper considers SDMU problems that can be modeled as Multi-stage
Stochastic Programs (MSPs). The modeling with MSPs help highlighting
how to use Lagrangian duality to derive exact and informative
gradients for parametric policies. Representing the uncertainty
explicit makes it easier to understand how TS-DDR avoids the implicit
function differentiation needed in related approaches. MSPs can be
represented by mathematical programs of the form:
\begin{align} \label{EQ: MSP}
    &\min_{(\mathbf{y}_1, \mathbf{x}_1) \in \mathcal{X}_1(\mathbf{x}_0)} f(\mathbf{x}_1, \mathbf{y}_1)  + \mathbf{E}[ \min_{(\mathbf{y}_2, \mathbf{x}_2) \in \mathcal{X}_2(\mathbf{x}_1, w_2)} f(\mathbf{x}_2, \mathbf{y}_2) + \\ 
    & \quad \quad \quad \mathbf{E}[\cdots +
     \mathbf{E}[\min_{(\mathbf{y}_t, \mathbf{x}_t) \in \mathcal{X}_t(\mathbf{x}_{t-1}, w_t)}f(\mathbf{x}_t, \mathbf{y}_t) + \mathbf{E}[\cdots]]] \nonumber
\end{align}
which minimizes a first stage cost function \( f(\mathbf{x}_1,
\mathbf{y}_1) \) and the expected value of future costs over possible
values of the exogenous stochastic variable \( \{w_{t}\}_{t=2}^{T} \in
\Omega \). Here, $\mathbf{x}_0$ is the initial system state and the
control decisions $\mathbf{y}_t$ are obtained at every period $t$
under a feasible region defined by the incoming state
$\mathbf{x}_{t-1}$ and the realized uncertainty \( w_t \).  This
optimization program assumes that the system is entirely defined by
the incoming state, a common modeling choice in many frameworks (e.g.,
MDPs \cite{murphy2000survey}). This is without loss of generality,
since any information can be appended in the state. The system
constraints can be generally posed as:
\begin{align} \label{EQ: MSP-Constraints}
    &\mathcal{X}_t(\mathbf{x}_{t-1}, w_t)= 
    \begin{cases}
        \mathcal{T}(\mathbf{x}_{t-1}, w_t, \mathbf{y}_t) = \mathbf{x}_t \\
        h(\mathbf{x}_t, \mathbf{y}_t) \geq 0 
    \end{cases}
\end{align}
where the outgoing state of the system $\mathbf{x}_t$ is a
transformation based on the incoming state, the realized uncertainty,
and the control variables. $h(\mathbf{x}_t, \mathbf{y}_t) \geq 0$
captures the state constraints. Markov Decision Process (MDPs) refer
to $\mathcal{T}$ as the ``transition kernel'' of the system. State and
control variables are restricted further by additional constraints
captured by $h(\mathbf{x}_t, \mathbf{y}_t) \geq 0$.  This paper
considers policies that map the past information into decisions. In
period $t$, an optimal policy is given by the solution of the dynamic
equations \cite{shapiro2009lectures}:
\begin{align} \label{EQ: DEQ-MSP}
    V_{t}(\mathbf{x}_{t-1}, w_t) &= \min_{\mathbf{x}_t, \mathbf{y}_t} & f(\mathbf{x}_t, \mathbf{y}_t) + \mathbf{E}[V_{t+1}(\mathbf{x}_t, w_{t+1})]  \quad \quad \\
    & \quad \quad \text{ s.t. } & \mathbf{x}_t  = \mathcal{T}(\mathbf{x}_{t-1}, w_t, \mathbf{y}_t) \nonumber  \quad \quad  \quad \quad  \quad \quad\\
    & & h(\mathbf{x}_t, \mathbf{y}_t)  \geq 0. \nonumber  \quad \quad  \quad \quad  \quad \quad  \quad \quad  \quad 
\end{align}


Since the state transition $\mathbf{x}_{t-1} \rightarrow
\mathbf{x}_{t}$ is deterministic once the uncertainty is observed, the
optimal policy can also be represented as a function of just the
realized uncertainties \( \{w_j\}_{j=2..t} \) and the initial state \(
\mathbf{x}_0 \):
\begin{align} 
& \mathbf{x}^{*}_t & & = \pi_t^{*}(\mathbf{x}^{*}_{t-1}, w_t) =  \pi_t^{*}(\mathbf{x}^{*}_{t-1}(\mathbf{x}^{*}_{t-2}, w_{t-1}), w_t) = \ldots \nonumber \\
& & & = \pi_t^{*}(\{w_j\}_{j=2..t}, \mathbf{x}_0) 
\end{align} 

\subsection{Value Iteration} \label{subsec: value_iter}

Value iteration approaches, such as SDDP and Q-learning, try to to
learn the value of an action in a particular state to inform
decisions. While Q-learning is a model-free algorithm that estimates
the state value by extensive evaluations, SDDP approximates the
expected future cost \(\mathbf{E}[V_{t+1}(\mathbf{x}_t, w_{t+1})]\)
by a piece-wise convex function calculating exact derivatives that
reduce the number of required evaluations. More precisely, SDDP
constructs the optimal policy by computing an outer approximation \(
\mathcal{V}_{t+1}(\mathbf{x}_t) \approx
\mathbf{E}[V_{t+1}(\mathbf{x}_t, w_{t+1})] \) and solving the
resulting problem:
\begin{align} \label{EQ: DEQ-V-MSP}
&     \pi_t^{*}(\{w_j\}_{j=2..t}, \mathbf{x}_0) \in & & \text{ arg}\min_{\mathbf{x}_t, \mathbf{y}_t} & & f(\mathbf{x}_t, \mathbf{y}_t) + \mathcal{V}_{t+1}(\mathbf{x}_t)  \nonumber \\
    & & & \text{ s.t. } & & \mathbf{x}_t  = \mathcal{T}(\mathbf{x}_{t-1}, w_t, \mathbf{y}_t)  \nonumber \\
    & & & & & h(\mathbf{x}_t, \mathbf{y}_t)  \geq 0  
\end{align}
\noindent
SDDP leverages the convexity of the problem and iteratively refines
the value function \( \mathcal{V}_{t+1}(\mathbf{x}_t)\) until
reaching optimality. Note that SDDP computes optimal decisions for the
first stage but its underlying ``policy'' requires solving of
optimization problems for all stages.

\subsection{Two-stage LDR} \label{subsec: ts_ldr}

Bodur et al \cite{bodur2022two} propose to learn linear policies for
the MSP defined in \eqref{EQ: MSP}. A linear policy approximates the
decision $\mathbf{x}_t$ at stage $t$ using 
\[
\sum_{j=2}^{t} \theta_{t,j} w_j + \theta_{t,1} \mathbf{x}_0
\]
where $\theta_{t,j}$ are the parameters to learn. These parameters can 
be obtained by solving a large-scale two-stage mathematical programming
problem of the form 
\begin{align}
    &\min_{\theta} \quad \mathbf{E}[\mathcal{Q}(w; \theta)] 
\end{align}
where the second stage $\mathbf{E}[Q(w; \theta)]$ computes the costs incurred by the policy for every possible realization of the stochastic process $w \in \Omega$ and $Q(w; \theta)$ is defined as follows:
\begin{align} \label{EQ: DR-MSP}
    &\min_{\mathbf{x}, \mathbf{y}, \delta} & &\sum_{t} f(\mathbf{x}_t, \mathbf{y}_t) + C^{\delta}_t||\delta_t|| \\
    & \; \; \text{ s.t. } & &\mathbf{x}_t = \mathcal{T}(\mathbf{x}_{t-1}, w_t, \mathbf{y}_t) \; &\forall t \nonumber \\
    & & & \mathbf{x}_t + \delta_t = \sum_{j=2}^{t} \theta_{t,j} w_j + \theta_{t,1} \mathbf{x}_0 \; :\lambda_t &\forall t \nonumber \\
    & & & h(\mathbf{x}_t, \mathbf{y}_t) \geq 0 \; \; &\forall t \nonumber 
\end{align}

In the second stage, the policy, $\sum_{j=2}^{t} \theta_{t,j} w_j +
\theta_{t,1} \mathbf{x}_0$, defines the target state to attain at
every period $t$, and the cost is a result of the optimal control
variable $\mathbf{y}$ and the slack variable $\delta$ that represents
target infeasibility. Adding the slack variable $\delta$ is a common
practice in stochastic programming to guarantee control feasibility
for any $\theta$, ensuring \textit{relatively complete recourse}
\cite{shapiro2009lectures}. $\lambda_t$ is the dual of the target
constraint, i.e., the derivative of the objective function with
respect to the target defined by the policy. If the second stage
\eqref{EQ: DR-MSP} is convex, the problem may be solved by benders
decomposition (or any similar method). Since the expectation is
typically computed through sampling, this approach typically suffers
from over-fitting; which can be mitigated using regularization
\cite{nazare2023solving},

\section{Two-Stage General Decision Rules} \label{sec: ts_gdr}

The first contribution of this paper is to generalize to TS-LDR to
Two-Stage General Decision Rules (TS-GDR), which supports arbitrary
decision rules.  In TS-GDR, the second stage problem $\mathcal{Q}(w; \theta)$ becomes
\begin{align} \label{EQ: TS-DDR-MSP}
  &\min_{\mathbf{x}, \mathbf{y}, \delta} & &\sum_{t} f(\mathbf{x}_t, \mathbf{y}_t) + C^{\delta}_t||\delta_t|| \nonumber\\
  & \text{ s.t. } & &\mathbf{x}_t = \mathcal{T}(\mathbf{x}_{t-1}, w_t, \mathbf{y}_t) \;         &\forall t \nonumber \\
  & & & \mathbf{x}_t + \delta_t = \pi_t(\{w_j\}_{j=2..t}, \mathbf{x}_0; \theta_t) \; :\lambda_t   &\forall t \nonumber \\
  & & & h(\mathbf{x}_t, \mathbf{y}_t) \geq 0 &\forall t \nonumber 
\end{align}
For specific parameter values, the target state is specified by the
policy, i.e., $\hat{\mathbf{x}}_t = \pi_t(\{w_j\}_{j=2..t},
\mathbf{x}_0; \theta_t)$, thus {\em decoupling the next state
  prediction from the optimization of decision variables $\mathbf{x},
  \mathbf{y}, \delta$}. It leads to the following equivalence:
$\mathcal{Q}(w; \theta) = Q(w, \hat{\mathbf{x}}_t) = $
\begin{align}
    &\min_{\mathbf{x}, \mathbf{y}, \delta} & &\sum_{t} f(\mathbf{x}_t, \mathbf{y}_t) + C^{\delta}_t||\delta_t||        \; \nonumber\\
    & \text{ s.t. } & &\mathbf{x}_t = \mathcal{T}(\mathbf{x}_{t-1}, w_t, \mathbf{y}_t)         & &\forall t \nonumber\\
    & & &\mathbf{x}_t + \delta_t = \hat{\mathbf{x}}_t  \;\; :\lambda_t  & &\forall t \nonumber \\
    & & & h(\mathbf{x}_t, \mathbf{y}_t) \geq 0   & &\forall t \nonumber
\end{align}
{\em Importantly, the second stage is a deterministic multi-period optimization.}

A policy can be any function $ \pi_t(\{w_j\}_{j=2..t}, \mathbf{x}_0;
\theta_t) $ parameterized by a set of parameters $\theta_t$. A TS-LDR
policy $\pi_t^l$ can be obtained by defining $s_t = [w_2, \cdots, w_t
  , \mathbf{x}_0]^\text{T}$ as the stacked history of uncertainty and
initial state:
\begin{align}
    \pi_t^l(\{w_j\}_{j=2..t}, \mathbf{x}_0; \theta_t) =\langle \theta_t, s_t \rangle.
\end{align}

The second contribution of this paper is to define Two-Stage Deep
Decision Rules (TS-DDR) where the policy $\pi_t^d$ is specified by a
deep neural network, i.e.,
\begin{align}
\pi_t(\{w_j\}_{j=2..t}, \mathbf{x}_0; \theta_t) &= W_K z_{K} + b_K \\
z_{n+1} &= \sigma ( W_n z_{n} + b_n) \; \forall n = 1 \ldots K-1 \nonumber\\
z_{1} &= \sigma ( W_0 s_t + b_0) \nonumber
\end{align}
where the $\theta_t = \{W_0, \dots, W_K, b_0, \dots b_k \}$ are the
parameters of a feed-forward deep neural network with \( K \) hidden
layers and nonlinear activation function \( \sigma \). Notice that a
function that just defines state variables $\mathbf{x}_t$ is also a
policy since the control variables $y_t$ are a direct consequence of
the state.

\section{Training of TS-DDR} \label{sec: ts_ddr}

This section presents the third contribution of the paper: the
training procedure for TS-DDR. While a TS-LDR policy can be trained by
solving a large-scale two-stage mathematical programming problem
(e.g., using Benders decomposition), the training of the TS-DDR uses a
combination of machine learning and mathematical programming
techniques. More specifically, {\it the training of the policy
  parameters uses deterministic optimization technology in the forward
  phase to satisfy the problem constraints and ensure feasibility, and
  stochastic gradient descent in the backward phase.} The link between
the two phases exploit {\em duality theory}.

At each iteration, the training procedure predicts the target state
$\hat{\mathbf{x}}_t$ and solves the second stage optimization. To
train the policy \( \pi \) parameterized by \( \theta \), the backward
phase differentiates through \( \mathcal{Q}(w; \theta) \)
and updates the policy parameters accordingly. Since the target
$\hat{\mathbf{x}}_t = \pi_t(\{w_j\}_{j=2..t}, \mathbf{x}_0; \theta_t)$
is a right-hand-side (rhs) parameter of the inner problem $Q$, by
duality theory, the subgradient of a subproblem objective with respect
to $\hat{\mathbf{x}}_t$ is given by the dual variable of its
associated constraint. If the problem is non-convex, the dual is the
subgradient of the local optimum. As a result,
\begin{align}
    \nabla_{\theta} \mathbf{E}[\mathcal{Q}&(w; \theta)] \approx \frac{1}{S}\sum_{s=1}^{S} \nabla_{\theta} \mathcal{Q}(w^s; \theta) \\
    &= \frac{1}{S} \sum_{s=1}^{S} \overbrace{\nabla_{\hat{\mathbf{x}}} Q(w^s, \hat{\mathbf{x}})}^{\lambda^s} \odot \nabla_{\theta} \pi(\{w^s_j\}_{j=2..T}, \mathbf{x}_0; \theta) \nonumber
\end{align}

where \( S \) is the number of samples.

By representing the policy in terms of only the realized
uncertainties, the proposed method can avoid the implicit function
differentiation of the KKT conditions, as done in
\cite{agrawal2019differentiable}, to back-propagate rewards through
the state transition towards the parametric policy. This is a major
advantage for applications with strict training times.

\paragraph{Time-Specific Policies} A time-specific policy \( \pi \) for TS-DDR can then be expressed as

\begin{align}
    \pi(\{w_j\}_{j=2..T}, \mathbf{x}_0; \theta) = \begin{bmatrix}
      \pi_t(w_1, \mathbf{x}_0; \theta_1) \\ \vdots
      \\ \pi_t(\{w_j\}_{j=2..T}, \mathbf{x}_0; \theta_T)
    \end{bmatrix}
\end{align}

and obtained using an automatic differentiation framework (e.g.,
Zygote.jl \cite{innes2019differentiable}) and a compatible optimizer
(e.g., Gradient Descent, Adam, AdaGrad) to update the policy
parameters.

\paragraph{Time-Invariant Policies}

An additional contribution of this paper with respect to TS-LDR is the
implementation of Time-Invariant Policies for TS-DDR. Indeed, TS-LDR
learns separate policies \( \pi_1, \ldots, \pi_T \) for each stage;
these policies have the signature \( \pi_t(\{w_j\}_{j=2..t},
\mathbf{x}_0; \theta) \), where \( \{w_j\}_{j=2..t} \) represents the
entire sequence of past uncertainties and \( \mathbf{x}_0 \) is the
initial state. This means each stage has a different number of inputs
since the history grows larger. TS-DDR uses $(\hat{\mathbf{x}}_t,
\ell_t) \leftarrow \pi(w_t^s, \ell_{t-1}; \theta_i)$, a single
time-invariant policy inspired by recurrent networks.  The policy
evaluation at each stage \( t \) shares a hidden (latent) state \(
\ell_t \) with the subsequent stage \( t+1 \). Trough training, \(
\ell_t \) carries the necessary information about \( \{w_j\}_{j=2..t}
\) and \( \mathbf{x}_0 \). Since the policy uses the same parameters
at every stage, it generalizes beyond the finite training horizon,
which is a distinct advantage. Indeed, it requires less computational
effort to train and fewer computational resources at execution time as
discussed in the next section.

\paragraph{The Training Algorithm}

\begin{algorithm}[!t]
\caption{TS-DDR Policy Estimation}\label{alg:training_alg}
\begin{algorithmic}[1]  
  \STATE \textbf{Input:} \(\pi(\cdot; \theta)\) \quad (\textit{initial policy parameters})
  \STATE Set initial parameters: \(\theta_0\)
  \FOR{\(i = 1\) to \(M\)}
    \STATE Set initial state conditions: \(\mathbf{x}_{0}\)
    \STATE Sample the stochastic process \(S\) times: 
           \(\{\{w_{t}^s\}_{t=1}^T\}_{s=1}^S\)
    \FOR{\(s = 1\) to \(S\)}
      \STATE Set initial latent state: \(\ell_0\)
      \STATE Compute first-stage target: 
        \[
          (\hat{\mathbf{x}}_1, \ell_1) \leftarrow \pi\bigl(\mathbf{x}_{0}, \ell_0; \theta_i\bigr)
        \]
      \STATE \(t \leftarrow 2\)
      \WHILE{\(t \leq T\)}
        \STATE Compute remaining targets:
        \[
          (\hat{\mathbf{x}}_t, \ell_t) \leftarrow \pi\bigl(w_t^s, \ell_{t-1}; \theta_i\bigr)
        \]
        \STATE \(t \leftarrow t + 1\)
      \ENDWHILE
      \STATE Solve the second-stage, implementation problem and record loss: 
             \(\textit{loss} \leftarrow Q\bigl(w^s; \hat{\mathbf{x}}\bigr)\)
      \STATE Compute the duals:
             \(\lambda^s \leftarrow \nabla_{\hat{\mathbf{x}}} Q\bigl(w^s, \hat{\mathbf{x}}\bigr)\)
      \STATE Compute the loss gradient:
             \[
               \nabla_{\theta_i}\mathcal{Q}\bigl(w^s; \theta_i\bigr) 
                 \leftarrow \lambda^s \;\odot\; \nabla_{\theta_i}\pi
             \]
    \ENDFOR
    \STATE Update parameters:
      \[
        \theta_{i+1} \leftarrow \theta_i 
        \;+\; \eta \left(\frac{1}{S}\sum_{s=1}^{S} \nabla_{\theta_i} \mathcal{Q}\bigl(w^s; \theta_i\bigr)\right)
      \]
  \ENDFOR
\end{algorithmic}
\end{algorithm}

The training algorithm for TS-DDR is shown in Algorithm
\ref{alg:training_alg}. Lines 6-13 are the forward pass: they compute
the target states (lines 6-12) before solving a second stage
subproblem through optimization. Lines 14-15 and line 17 are the
backward pass. It is interesting to highlight the similarities and
differences between RL and TS-DDR. Both share a forward phase
evaluating various scenarios with decisions over time and an update of
the policy parameters through gradient computations. However, TS-DDR
uses dedicated optimization technology to minimize costs and ensure
feasibility in its forward phase, as well as exact gradients obtained
through the nature of the optimizations and duality theory. {\em This
  integration of RL and multi-stage stochastic optimization is a key
  contribution of this paper for solving MSPs.}



\section{Decision Making process} \label{sec: decision_making}


To make first-stage decisions, SDDP (a time-specific method) iterates
over scenarios to approximate future outcomes. It considers future
decisions just enough to provide some guarantees regarding future
costs. This process does not ensure that future policies have
converged to their optimal state. Moreover, future periods are likely
to encounter states not simulated during training. As a result,
decision makers must rerun the SDDP procedure over time, even before
the initial training horizon ends, which is highly time-consuming.
Time-specific models, such as Two-Stage LDR, may encounter similar
issues since they do not account for all possible scenarios. However,
this is effectively mitigated by training under appropriate model
assumptions and using regularization, as demonstrated in
\cite{nazare2023solving}. Time-specific models also require retraining
at the end of the considered time horizon.  TS-GDR utilizes
regularization via SGD and employs a typical
``training-validation-test'' split process to check for
convergence. Additionally, with time-invariant policies, TS-DDR avoids
the finite horizon limitation, further reducing the need for
retraining.

\section{Long-Term Hydrothermal Dispatching} \label{sec: hyd}

\paragraph{Problem Description:}
Energy storage is a cornerstone in the quest for sustainable energy
solutions, offering a critical avenue for emission reduction and grid
optimization. Their dispatchable nature mitigates the intermittency of
renewable energy sources like wind and solar, thus enhancing grid
stability and enabling greater reliance on clean energy. Among the
different types of storage, hydro reservoirs are prominent large-scale
energy storage systems. These reservoirs are capable of efficiently
storing and releasing vast amounts of energy. Their mechanism, that
involves the storing of water during periods of low demand and the
subsequent releases to generate electricity during peak hours,
provides a critical means of balancing supply and demand.

The optimization of sequential decision strategies (policies) to
manage energy storage systems and minimize energy dispatch costs is
referred to as the LTHD problem. The LTHD is a complicated MSP problem
as it needs to take into account the complex power flow equations that
represent the physics of electricity transmission.  The non-convex
nature of AC power flow inhibits the use of classical optimization
methods such as SDDP. Moreover, other control and ML methods (such as
RL) have difficulties scaling with decision and scenario
dimensions. It is an important case study to evaluate the benefits of
any novel method. The second-stage policy implementation 
is described in Appendix \ref{Annex:LTHD}.

\paragraph{Experimental Setting:} {\em The results are reported for linear, conic, and nonlinear
  nonconvex problems to consider fundamental different problem
  structures.} These are obtained from the DCLL approximation of power
systems (linear), the Second-Order Cone (SOC) relaxation (conic), and
the AC-PF (non-convex). See Appendix \ref{annex: DCLL} and \ref{annex:
  SOC} for the descriptions of linear and conic cases. Appendix
\ref{annex: extra_results_bolivia} provides the computational
resources and the hyperparameters used.

The evaluation compares the time-invariant TS-DDR, SDDP, and TS-LDR.
For the linear and conic case, SDDP is optimal for stage-independent
uncertainty, which allows for estimating how accurate TS-DDR is. The
evaluation also compares the inconsistent policies trained under SOC
and DCLL with TS-DDR. The evaluation also compares TS-DDR with TS-LDR
\cite{bodur2022two}. However, instead of using explicit regularization
to avoid over-fitting, as in \cite{nazare2023solving}, the
implementation uses SGD.

\paragraph{The 28-bus Case Study from Bolivia:}
To examine the scalability of TS-DDR under different power flow
formulations, the results employ a realistic case study based on the
Bolivian power system.  Results on a smaller 3-bus system are shown in
the Appendix for completeness with prior results. The Bolivian system
comprises 28 buses, 26 loads, 34 generators (including 11 hydro
units), and 31 branches, with a predominantly radial configuration and
only three loops. The planning horizon extends to 96 periods. For each
stage, the evaluation uses 165 scenarios from historical data.

\paragraph{Results:}
Tables \ref{tab:bolivia_dcll}-\ref{tab:bolivia_ac} shows the
out-of-sample test results for different policies in the LTHD problem
instances. Tables have 5 columns: (1) "Model" signaling the policy
type; (2) "Plan" representing the power flow formulation used in
training; (3) "Imp Costs" meaning the average grid operational costs
under the power flow formulation described in the table caption ($\pm$
the standard deviation); (4) "GAP" making explicit the average
percentage difference of the models' \textit{Imp Costs} with the best
model ($\pm$ the standard deviation); (5) "Training" presenting the
amount of time in training the policy; and (6) "Execution" shows the
amount of time need for producing implementable policies. For each
experiment (``Case \& Formulation''), policies are ordered by their
implementation costs.

In the convex cases (Tables
\ref{tab:bolivia_dcll}-\ref{tab:bolivia_soc}), the SDDP Implementation
(Imp) cost serves as the reference point for the calculation of the
optimality GAP. However, as previously discussed, when the planning
formulation differs from the implementation one, and the resulting
SDDP policy is time-inconsistent and thus sub-optimal. The GAP, in
Table\ref{tab:bolivia_ac}, is then calculated with respect to the best
performing model.

By not (over) fitting to the finite horizon of the experiments, the
TS-DDR policy is sub-optimal, which explains its non-zero optimality
GAP. In fact, the TS-DDR polices have higher implementation costs than
other benchmarks in both the DCLL and AC formulations for the 3-bus
case. Nevertheless, the optimality to the best found policy
implementation cost decreases as the problem becomes bigger and the
assumptions of the remaining benchmarks are increasingly violated. For
the Bolivian case under the AC (non-convex) formulation, the TS-DDR
policy is the best performing policy beating all other policies in
both terms of training computational resources and implementation
costs.

More precisely, on the convex cases, TS-DDR is within 0.63\% and
0.49\% of optimality and is 4 orders of magnitude faster than SDDP at
execution time. Interestingly, even the training time of TS-DDR is an
order of magnitude faster than the SDDP execution times, On the
non-convex case, TS-DDR outperforms SDDP by at least 0.32\% in
solution quality, and brings orders of magnitude improvements in
solution times again. TS-DDR also produces order of magnitude
improvements in solution quality compared to TS-LDR, highlighting the
benefits of complex policy representations.

\begin{table*}[!t]
\centering
\caption{Comparison of SDDP and ML Decision Rule for Bolivia with DCLL Implementations.}
\begin{tabular}{ccccccc}
\toprule
\textbf{Model} & \textbf{Plan} & \textbf{Imp Cost (USD)} & \textbf{GAP (\%)} & \textbf{Training (Min)} & \textbf{Execution (Min)} \\
\midrule
SDDP & DCLL & $295879 (\pm 5667)$ & - & - & 10 \\
TS-DDR & DCLL & $297757 (\pm 5169)$ & $0.63 (\pm 2.71)$ & 2.5 & 0.002 \\
TS-LDR & DCLL & $300806 (\pm 8054)$ & $1.66 (\pm 0.72)$ & 15 & 0.002 \\
\bottomrule
\end{tabular}
\label{tab:bolivia_dcll}
\end{table*}

\begin{table*}[!t]
\centering
\caption{Comparison of SDDP and ML Decision Rule for Bolivia with SOC Implementations.}
\begin{tabular}{ccccccc}
\toprule
\textbf{Model} & \textbf{Plan} & \textbf{Imp Cost (USD)} & \textbf{GAP (\%)} & \textbf{Training (Min)} & \textbf{Execution (Min)} \\
\midrule
SDDP & SOC & $300219 (\pm 5176)$ & - & - & 480 \\
TS-DDR & SOC & $301694 (\pm 4856)$ & $0.49 (\pm 2.38)$ & 40 & 0.025 \\
TS-LDR & SOC & $313205 (\pm 5295)$ & $4.32 (\pm 2.42)$ & 150 & 0.025 \\
\bottomrule
\end{tabular}
\label{tab:bolivia_soc}
\end{table*}

\begin{table*}[!t]
\centering
\caption{Comparison of SDDP and ML Decision Rule for Bolivia with AC Implementation.}
\begin{tabular}{ccccccc}
\toprule
\textbf{Model} & \textbf{Plan} & \textbf{Imp Cost (USD)} & \textbf{GAP (\%)} & \textbf{Training (Min)} & \textbf{Execution (Min)} \\
\midrule
TS-DDR & AC & $301851 (\pm 4876)$ & - & 60 & 0.067 \\
SDDP & SOC & $302816 (\pm 5431)$ & $0.32 (\pm 2.34)$ & - & 480 \\
TS-LDR & AC & $319326 (\pm 4715)$ & $5.79 (\pm 1.30)$ & 226.01 & 0.067 \\
SDDP & DCLL & $323895 (\pm 3944)$ & $7.30 (\pm 2.27)$ & - & 320 \\
\bottomrule
\end{tabular}
\label{tab:bolivia_ac}
\end{table*}

\paragraph{Reinforcement Learning:}

\begin{figure}[!t]
    \centering
    \includegraphics[width=0.45\textwidth]{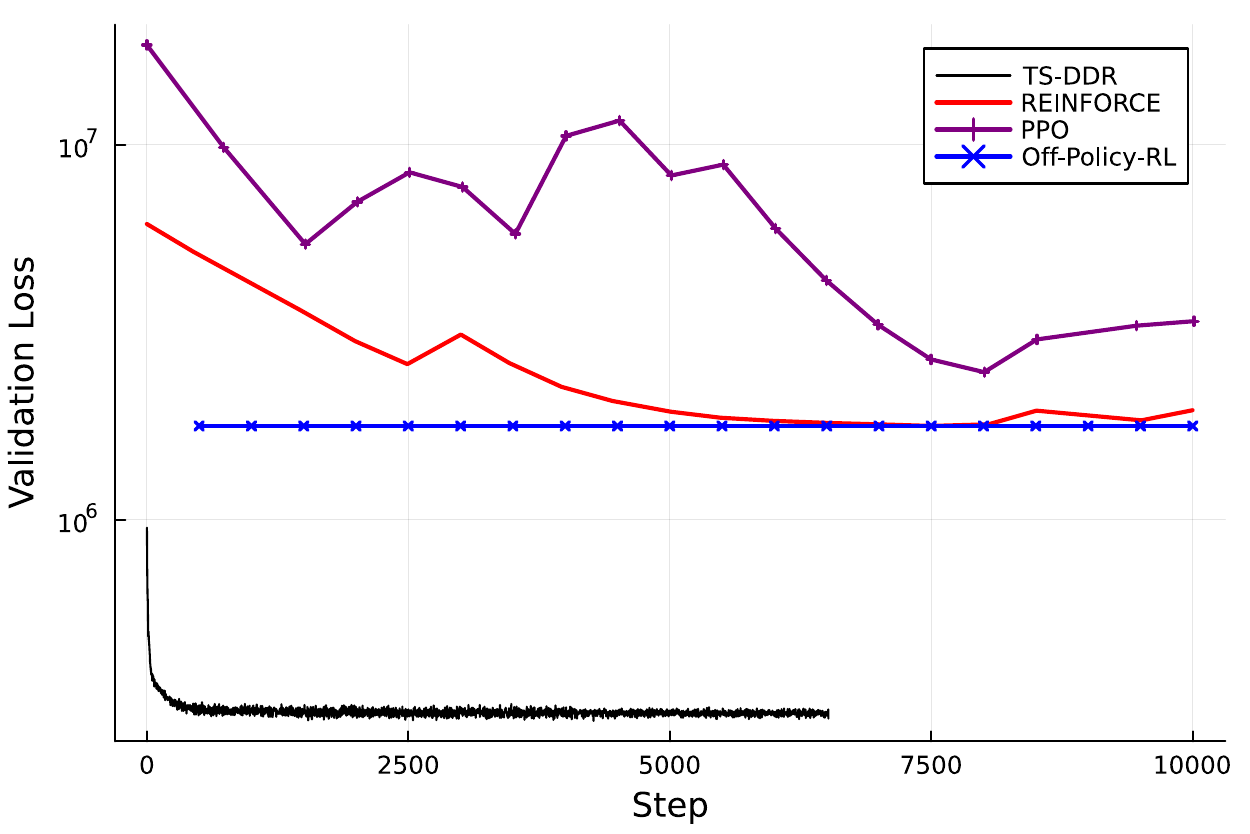}
    \caption{LTHD Training Curves for Model-Free RL Methods and TS-GDR Baseline}
    \label{fig:hydro_benchmark}
\end{figure}

In order to assess the advantage of using TS-DDR compared to
traditional RL methods, the evaluation considers the application of
standard algorithms in the RL literature to LTHD. Appendix
\ref{annex: RL} provides a description of these algorithms.  Figure
\ref{fig:hydro_benchmark} highlights that some methods showed
promising initial results with cumulative rewards increasing over time
(i.e., a decrease in operating costs). However, the improvements then
level off and the methods stagger at a very low quality solution.  Moreover,
no RL model-free method was able to find feasible policies - i.e. in
which there is no discrepancy between the target state
$\hat{\mathbf{x}}_t$ and the achieved state $\mathbf{x}_t$. These
results highlight the challenges documented in constrained learning
spaces such as those encountered in training Physics-Informed Neural
Networks (PINNs). As noted by \cite{liu2024config}, learning problems
involving multiple additive terms with conflicting objectives, like
penalties for satisfying various constraints, leads to gradients with
conflicting update directions. This suggests that achieving better
results with model-free RL for the LTHD may require careful tuning of
constraint penalties and a more balanced trade-off with the reward
function, as seen in approaches like
\cite{xuan2023constrained}. Nevertheless, no successful RL approach
has been proposed for the general case. This paper proposes TS-DDR as
a model-based constrained RL approach that can be used along side
other important techniques from the RL literature to overcome other
issues such as epistemic uncertainty.

\paragraph{Storage Behavior:}

\begin{figure}[!t]
    \begin{minipage}[t]{0.4\textwidth}
        \centering
        \includegraphics[width=\linewidth]{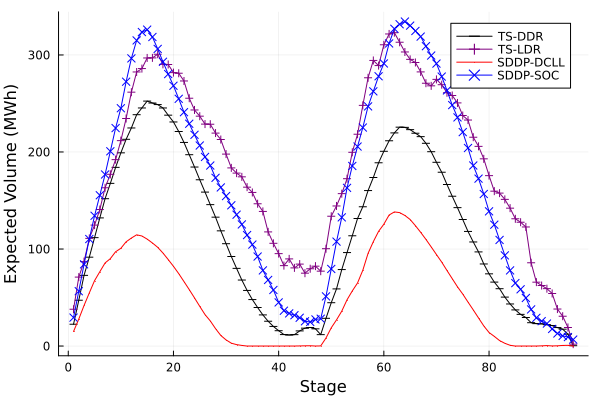}
    \end{minipage}
    \caption{Expected Stored Energy for the AC Formulation.}
    \label{fig:comp-ac}
\end{figure}

Figure \ref{fig:comp-ac} shows the expected stored energy (volume) and
thermal generators dispatch over time for each analysed policy in the
Bolivian grid under the AC power flow formulation.  TS-DDR stores less
water over time compared to most of the other policies. It is capable
of finding a cost effective strategy that avoids more expensive
generators. SDDP polices, in contrast, tend to exhibit higher needs
for stored energy. This is in line with the findings in
\cite{rosemberg2021assessing}. 

\section{The Goddard Rocket Control Problem} \label{annex: mpc}

To showcase the capability of TS-DDR to learn other complex policies,
this section considers the Goddard Rocket control (GRC) problem from
``Benchmarking Optimization Software with COPS 3.0''. The evaluation
considers 1,200 stages and the test case was modified to have a random
force, $w_t \in \mathcal{N}(0,1$, affecting the velocity transition equation, i.e.,
\begin{align}
& \frac{v_t - v_{t-1}}{\Delta t} = \frac{u_{t-1} - D(h_{t-1}, v_{t-1})}{m_{t-1}} - g(h_{t-1}) \overbrace{- w_{t-1}} \nonumber
\end{align}

While SDDP is a stochastic
version of the common Model Predictive Control (MPC) approach, it requires convexity and, for non-convex
applications, convex relaxations and approximations need be used for estimating
the value functions. However, for the GRC problem, no such convex approximation is readily available, therefore the fallback is to solve the deterministic version
of the multi-stage problem and update initial conditions based on the realized
uncertainty in a rolling horizon fashion. This is suboptimal in
general and, for the GRC problem, the gap with
TS-DDR is $23\%$. Table \ref{table:goddard} also highlights the
significant computational benefits of TS-DDR.

\begin{table}[!t]
\caption{Comparison of MPC and TS-DDR for the Goddard Case.}
\centering
\begin{tabular}{|l|c|c|c|}
\hline
\textbf{Model} & \textbf{Training} & \textbf{Inference} & \textbf{GAP} $\%$ \\
\hline
\textbf{TS-DDR} & 35 min.  & 0.091 sec. & $-$ \\
\textbf{MPC} & None & 2 hours & $23.42(\pm 12.62)$ \\
\hline
\end{tabular}
\label{table:goddard}
\end{table}

\section{Limitations}
\label{sec: limitations}

\begin{figure}[!t]
    \begin{minipage}[t]{0.4\textwidth}
        \centering
        \includegraphics[width=\linewidth]{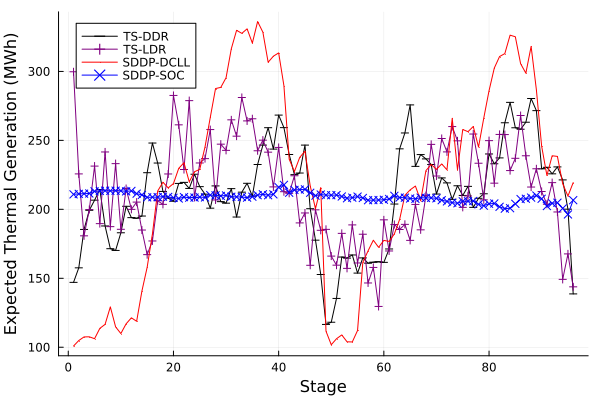}
    \end{minipage}
    \caption{Expected Thermal Dispatch for the AC Formulation.}
    \label{fig:comp-ac-2}
\end{figure}

There are a number of limitations in this study, as well as additional
opportunities. First, TS-GDR is a general framework and the paper only
considers one of its instantiations: it would be interesting to
explore other machine learning models. Second, the LTHD problem is
particularly difficult for RL due to its complexity and high
dimensionality. It would be interesting however to study how TS-GDR
behaves on applications where RL has been effective, to gain deeper
insights into each approach's strengths and weaknesses. Third, the
scalability of TS-DDR on extremely large instances needs to be
investigated. This paper focuses on an application for which SDDP was
applicable. Evaluating TS-DDR for applications outside the reach of
SDDP is an important direction. Fourth, it would be interesting to
evaluate TS-DDR on some additional applications where its strengths
can be highlighted. There is a scarcity of test cases in this domain
and the team aims at addressing this limitation. Figure
\ref{fig:comp-ac-2} highlights another source of possible improvement
for the two stage policies, TS-DDR and TS-LDR, which exhibit a larger
variance in the amount of thermal generation dispatch since, in some
applications, control actuators might have frequency
restrictions. This could be solved by either modeling these
restrictions explicitly or by adding regularization penalties for
intermittent behaviour. Appendix \ref{annex: extra_results_bolivia}
shows the same pictures for both the SOC and DCLL formulations.

\section{Conclusion} \label{sec: conclusion}

This paper introduced TS-GDR and its instantiation TS-DDR, a novel
approach that integrates machine learning with stochastic optimization
to solve Multistage Stochastic Optimization Problems. TS-DDR can be
trained effectively by solving deterministic optimization problems in
the forward pass and using closed-form gradients in the backward pass
thanks to duality theory. At inference time, TS-DDR relies on the
evaluation of a deep learning network. was validated on the Long-Term
Hydrothermal Dispatch Problem (LTHD) and actual instances for Bolivia,
a problem of significant societal impact, as well as the Goddard
rocket problem. The results demonstrated the effectiveness of TS-DDR
in improving solution quality while significantly reducing computation
times.

TS-DDR potentially represents a significant step forward in sequential
decision-making under uncertainty.  By leveraging the complementary
strengths of machine learning and stochastic optimization, TS-DDR
offers a versatile and scalable framework. Investigating other hybrid
approaches merging RL and stochastic optimization is certainly a
promising direction.  Other opportunities include the study of
alternative parametric families for policies, applying TS-DDR in other
domains, addressing the challenges posed by discrete variables, and
alternative risk measures.




\bibliography{bibly}
\bibliographystyle{IEEEtran}

\appendix
\onecolumn

\subsection{LTHD Policy Implementation Problem} \label{Annex:LTHD}

\begin{subequations}
\label{eq:LTHD}
\footnotesize
\begin{align}
    Q\bigl(w; \hat{\mathbf{x}}\bigr) 
    &= \min_{\mathbf{x}, \mathbf{y}, \delta} 
       \sum_{t} \sum_{i \in \mathcal{I}} 
         C_{it} \, p^{g}_{it} 
       \;+\; C^{\delta}_t \,\bigl\|\delta_t\bigr\|
       \label{eq:LTHD:cost}
       \\
    &\text{s.t.}\quad 
      \mathbf{x}_{jt} + u_{jt} + s_{jt} 
      \;=\;
      \mathbf{x}_{j,t-1} + A_{j,t}(\omega_t) \nonumber \\
      & \;+\; \sum_{k \in \mathcal{H}^U_j} u_{kt} 
      \;+\; \sum_{k \in \mathcal{H}^{S}_j} s_{kt},
      \quad \forall\, t,\; j \in \mathcal{H},
      \label{eq:LTHD:hydro}
      \\
    &\quad\quad
      u_{jt} = \Phi_j \, p^{g}_{jt},
      \quad \forall\, t,\; j \in \mathcal{H},
      \label{eq:LTHD:production}
      \\
    &\quad\quad
      \mathbf{x}_t + \delta_t \;=\; \hat{\mathbf{x}}_t 
      \quad :\lambda_t,
      \quad \forall\, t,
      \label{eq:LTHD:larget}
      \\
    &\quad\quad
      p^{g}_t \;\in\; \text{PF}_t,
      \quad \forall\, t.
      \label{eq:LTHD:powerflow}
\end{align}
\end{subequations}

The second-stage policy implementation for the LTHD problem is described in Equation \ref{eq:LTHD}. The objective \eqref{eq:LTHD:cost} is to minimize the total cost of generation dispatch while meeting electricity demand and adhering to physical and engineering constraints. State variables $\mathbf{x}$ represent the volumes of the hydro reservoirs, while control variables $u$ denote the outflows from these reservoirs, and $\pg$, the energy production variables for all generators (not only hydro). $\delta$ are the deviation variables ensuring that the state meets the target. The model considers uncertain scenarios through $\omega_t$, which is accounted in the stochastic inflow of water to the reservoirs, $A_{j,t}(\omega_t)$. $s_{jt}$ represents the spillage, which is the excess water that cannot be stored and must be released.

The key constraints of the model include: \eqref{eq:LTHD:hydro} which ensures the balance of water volumes in the reservoirs by accounting for inflows ($A_{j,t}(\omega_t)$), outflows ($u_{jt}$), spillage ($s_{jt}$), and the flow ($u_{kt},s_{kt}$) from upstream reservoirs ($\mathcal{H}^U, \mathcal{H}^{S}$). \eqref{eq:LTHD:production} relates the outflows to energy production via the production factor $\Phi_j$. \eqref{eq:LTHD:larget} ensures that the state and deviation $\delta_t$ match the target $\hat{\mathbf{x}}_t$, with $\lambda_t$ representing the dual variables for these constraints. Finally, \eqref{eq:LTHD:powerflow} ensures that the energy dispatch respects power flow equations and generator limits defined by the set $\text{PF}_t$ for each stage $t$.

The most accurate formulation of the physical constraints, $\text{PF}_t$, were formalized in \cite{carpentier1962contribution} to describe AC power-flow (AC-PF). Annex \ref{annex: AC} provides a general overview of these constraints. The AC-PF model is a non-convex non-linear problem (NLP), not suitable for the classical SDDP algorithm. Thus, as in many applications, convex approximations and relaxations can be used to meet the SDDP convexity assumption \cite{molzahn2019survey}. Specifically, planning agents utilize simplified models to compute cost-to-go functions and couple them to optimization problems that guarantee a feasible operative decision.\footnote{Annex \ref{annex: sddp} details the coupling of planning cost-to-go functions to find implementable and how to simulate similar polices efficiently.}

\subsection{Alternating Current (AC) - Power Flow} \label{annex: AC}

This appendix details the formulation of the Alternating Current Power Flow (AC-PF) problem, which is fundamental for analyzing and optimizing electrical power systems. The AC-PF model captures the complex relationships between voltages, power generation, and power consumption across the network, ensuring adherence to physical laws and operational constraints. The following equations (\ref{eq:ACPF}) provide a comprehensive mathematical representation of these relationships, incorporating key constraints such as Kirchhoff's current law, Ohm's law, thermal limits, and bounds on voltage and power generation. By attending the AC-PF constraints, we ensure a feasible dispatch of power generation to meet demand while maintaining system stability and efficiency. For clarity, certain complexities such as transformer tap ratios, phase angle difference constraints, and reference voltage constraints are not included in this presentation.

\begin{subequations}
\label{eq:ACPF}
\footnotesize
    \begin{align}
    \text{PF} = \Big\{ \pg \; | \; 
            & \Sg_{i} - \Sd_{i} - (Y^{s}_{i})^{\star} |\V_{i}|^{2} = \sum_{ij \in \mathcal{E} \cup \mathcal{E}^{R}} \Sf_{ij}
            && \forall i \in \mathcal{N} 
            \label{eq:ACPF:kirchhoff} \\
        & \Sf_{ij} = (Y_{ij} + Y_{ij}^{c})^{\star} |\V_{i}|^{2} - Y_{ij}^{\star} \V_{i} \V_{j}^{\star}
            && \forall ij \in \mathcal{E} \label{eq:ACPF:ohm_fr}\\
        & \Sf_{ji} = (Y_{ij} + Y_{ji}^{c})^{\star} |\V_{j}|^{2} - Y_{ij}^{\star} \V_{i}^{\star} \V_{j}
            && \forall ij \in \mathcal{E} \label{eq:ACPF:ohm_to}\\
        & |\Sf_{ij}|, |\Sf_{ji}| \leq \bar{s}_{ij}
            && \forall ij \in \mathcal{E} \label{eq:ACPF:thermal_limits} \\
        & \vmmin_{i} \leq |\V_{i}| \leq \vmmax_{i} 
            && \forall i \in \mathcal{N} \label{eq:ACPF:voltage_bounds}\\
        & \pgmin_{i} \leq \pg_{i} \leq \pgmax_{i}
            && \forall i \in \mathcal{N} \label{eq:ACPF:active_dispatch_bounds}\\
        & \qgmin_{i} \leq \qg_{i} \leq \qgmax_{i} 
            && \forall i \in \mathcal{N} \label{eq:ACPF:reactive_dispatch_bounds}
            \Big\}
    \end{align}
\end{subequations}

Equation \ref{eq:ACPF} presents the AC-PF formulation, in complex variables.
Constraints \eqref{eq:ACPF:kirchhoff} enforce power balance (Kirchhoff's current law) at each bus.
Constraints \eqref{eq:ACPF:ohm_fr} and \eqref{eq:ACPF:ohm_to} express Ohm's law on forward and reverse power flows, respectively.
Constraints \eqref{eq:ACPF:thermal_limits} enforce thermal limits on forward and reverse power flows.
Finally, constraints \eqref{eq:ACPF:voltage_bounds}--\eqref{eq:ACPF:reactive_dispatch_bounds} enforce minimum and maximum limits on nodal voltage magnitude, active generation, and reactive generation, respectively.

\subsection{Second Order Cone (SOC) - Power Flow} \label{annex: SOC}

This appendix introduces the Second-Order Cone Power Flow (SOC-PF) formulation, a convex relaxation of the AC-PF model. The SOC-PF model simplifies the problem by introducing additional variables and relaxing certain non-convex constraints, resulting in a more tractable problem while maintaining a close approximation to the original AC-PF. The following equations (\ref{eq:SOCPF}) encapsulate the SOC-PF formulation, addressing active and reactive power balance, Ohm's law, thermal limits, voltage bounds, and generation constraints.

\begin{subequations}
    \label{eq:SOCPF}
    \footnotesize
        \begin{align}
        \text{PF} = \Big\{ \pg \; | \; 
            & \pg_{i} - \pd_{i} - g^{s}_{i} \wmsoc_{i} = \sum_{ij \in \mathcal{E} \cup \mathcal{E}^{R}} \pf_{ij}
                && \forall i \in \mathcal{N} 
                \label{eq:SOCPF:kirchhoff_active} \\
            & \qg_{i} - \qd_{i} + b^{s}_{i} \wmsoc_{i} = \sum_{ij \in \mathcal{E} \cup \mathcal{E}^{R}} \qf_{ij}
                && \forall i \in \mathcal{N} 
                \label{eq:SOCPF:kirchhoff_reactive} \\
            & \pf_{ij} = \gamma^{p}_{ij} \wmsoc_{i} + \gamma^{p,r}_{ij} \wrsoc_{ij} + \gamma^{p,i}_{ij} \wisoc_{ij}
                && \forall ij \in \mathcal{E} 
                \label{eq:SOCPF:ohm:active:fr}\\
            & \qf_{ij} = \gamma^{q}_{ij} \wmsoc_{j} + \gamma^{q,r}_{ij} \wrsoc_{ij} + \gamma^{q,i}_{ij} \wisoc_{ij}
                && \forall ij \in \mathcal{E} 
                \label{eq:SOCPF:ohm:reactive:fr}\\
            & \pf_{ji} = \gamma^{p}_{ji} \wmsoc_{i} + \gamma^{p,r}_{ji} \wrsoc_{ij} + \gamma^{p,i}_{ji} \wisoc_{ij}
                && \forall ij \in \mathcal{E}
                \label{eq:SOCPF:ohm:active:to}\\
            & \qf_{ji} = \gamma^{q}_{ji} \wmsoc_{j} + \gamma^{q,r}_{ji} \wrsoc_{ij} + \gamma^{q,i}_{ji} \wisoc_{ij}
                && \forall ij \in \mathcal{E}
                \label{eq:SOCPF:ohm:reactive:to}\\
            & (\pf_{ij})^{2} + (\qf_{ij})^{2} \leq \bar{s}_{ij}^{2}
                && \forall ij \in \mathcal{E} \cup \mathcal{E}^{R} 
                \label{eq:SOCPF:thermal_limits} \\
            & (\wrsoc_{ij})^{2} + (\wisoc_{ij})^{2} \leq \wmsoc_{i} \wmsoc_{j}
                && \forall ij \in \mathcal{E}
                \label{eq:SOCPF:jabr}\\
            & \vmmin_{i}^{2} \leq \wmsoc_{i} \leq \vmmax_{i}^{2} 
                && \forall i \in \mathcal{N} \label{eq:SOCPF:voltage_bounds}\\
            & \pgmin_{i} \leq \pg_{i} \leq \pgmax_{i}
                && \forall i \in \mathcal{N} \label{eq:SOCPF:active_dispatch_bounds}\\
            & \qgmin_{i} \leq \qg_{i} \leq \qgmax_{i} 
                && \forall i \in \mathcal{N} \label{eq:SOCPF:reactive_dispatch_bounds}
                \Big\}
        \end{align}
\end{subequations}

The SOC-PF formulation introduces additional variables:
\begin{align}
    \wmsoc_{i} &= \vm_{i}^{2}, 
        && \forall i \in \mathcal{N}\\
    \wrsoc_{ij} &= \vm_{i} \vm_{j} \cos (\theta_{j} - \theta_{i}), 
        && \forall ij \in \mathcal{E}\\
    \wisoc_{ij} &= \vm_{i} \vm_{j} \sin(\theta_{j} - \theta_{i}), 
        && \forall ij \in \mathcal{E}
\end{align}

The non-convex constraint:
\begin{align}
    \label{eq:SOC:quad_prod}
    (\wrsoc_{ij})^{2} + (\wisoc_{ij})^{2} &= \wmsoc_{i} \wmsoc_{j}, && \forall ij \in \mathcal{E}
\end{align}
is relaxed to:
\begin{align}
    \label{eq:SOC:Jabr}
    (\wrsoc_{ij})^{2} + (\wisoc_{ij})^{2} &\leq \wmsoc_{i} \wmsoc_{j}, && \forall ij \in \mathcal{E}
\end{align}

Equation \ref{eq:SOCPF} presents the SOC-PF formulation using real variables. Constraints \eqref{eq:SOCPF:kirchhoff_active} and \eqref{eq:SOCPF:kirchhoff_reactive} enforce Kirchhoff's current law for active and reactive power at each node. Constraints \eqref{eq:SOCPF:ohm:active:fr}--\eqref{eq:SOCPF:ohm:active:to} capture Ohm's law on active and reactive power 
flows. The $\gamma$ parameters derive from substituting variables $\wmsoc, \wrsoc, \wisoc$ in \eqref{eq:ACPF:ohm_fr}--\eqref{eq:ACPF:ohm_to}. Constraints \eqref{eq:SOCPF:thermal_limits} enforce thermal limits on power flows. Constraint \eqref{eq:SOCPF:jabr} is Jabr's inequality. Finally, constraints \eqref{eq:SOCPF:voltage_bounds}--\eqref{eq:SOCPF:reactive_dispatch_bounds} enforce limits on nodal voltage magnitude, active, and reactive generation.

The SOC-PF formulation is nonlinear and convex, making it more tractable than AC-PF and solvable using polynomial-time interior-point algorithms. As a relaxation of AC-PF, SOC-PF provides valid dual bounds on the optimal value of AC-PF.

\subsection{Direct Current with Line Losses (DCLL) - Power Flow} \label{annex: DCLL}

This appendix introduces the Direct Current with Line Losses Power Flow (DCLL-PF) formulation, a quadratic approximation of AC-PF. This approximation assumes all voltage magnitudes are one per-unit, voltage angles are small, losses are quadratically proportional to the flow, and reactive power is ignored. The DCLL-PF improves on the DC-PF by approximating line losses, making it a more accurate linear approximation widely used in electricity markets and planning problems.

\begin{subequations}
    \label{eq:DCPF}
    \footnotesize
    \begin{align}
        \text{PF} = \Big\{ \pg \; | \; 
            & \pg_{i} + \sum_{ji \in \mathcal{E}} \pf_{ji} - \sum_{ij \in \mathcal{E}} \pf_{ij}  = \pd_{i}
                && \forall i \in \mathcal{N}
                \label{eq:DCPF:power_balance}\\
            & \pf_{ij} = b_{ij} (\va_{j} - \va_{i}) 
                && \forall ij \in \mathcal{E}
                \label{eq:DCPF:ohm}\\
            & \pf_{ij} + \pf_{ji} \geq 
                \frac{g_{ij}}{g_{ij}^2 + b_{ij}^2}\pf_{ij}^2 
                && \forall ij \in \mathcal{E} \label{eq:DCPF:losses}\\
            & |\pf_{ij}| \leq \bar{s}_{ij}
                && \forall ij \in \mathcal{E}
                \label{eq:DCPF:bounds:pf}\\
            & \pgmin_{i} \leq \pg_{i} \leq \pgmax_{i} 
                && \forall i \in \mathcal{N}
                \label{eq:DCPF:bounds:pg}
         \Big\}
    \end{align}
    \end{subequations}

Equation \ref{eq:DCPF} presents the DCLL-PF formulation using a linear programming (LP) approach. Constraints \eqref{eq:DCPF:power_balance} enforce active power balance at each node. Constraints \eqref{eq:DCPF:ohm} approximate Ohm's law using a phase-angle formulation. Constraints \eqref{eq:DCPF:losses} ensure the consideration of line losses. Constraints \eqref{eq:DCPF:bounds:pf} enforce thermal constraints on each branch. Constraints \eqref{eq:DCPF:bounds:pg} enforce limits on active power generation. Constraints on phase angle differences and slack bus are omitted for readability but are implemented in numerical experiments.

\subsection{Implementable Decisions in Stochastic Dual Dynamic Programming (SDDP)} \label{annex: sddp}

This appendix details how decision-makers use convex models to get implementable decisions for non-convex systems.

In SDDP, planning agents utilize simplified (convex) models to compute cost-to-go functions, which guide decision-making over medium- and long-term horizons. These functions provide insights into the optimal reservoir levels and system operation strategies. However, for operational implementation, it is essential to ensure that the derived decisions align with the intricacies of the real network.

To translate planning decisions into operational actions, Independent System Operators (ISOs) employ a coupling approach. Initially, the SDDP algorithm converges using simplified convex network models. Subsequently, the second-stage cost-to-go function is integrated into more detailed models that offer a realistic portrayal of the system.

A common strategy employed by ISOs involves a rolling-horizon operating scheme. In this approach, decisions obtained from the planning stage, embedded in the cost-to-go function, are periodically updated with real-time data. For instance, in regions like Brazil and Chile, ISOs update the state variables, such as reservoir levels, and re-converge SDDP in subsequent periods.

However, fitting a policy under a simplified convex model but implementing decisions in the detailed non-convex reality produces what is called time-inconsistent (sub-optimal) policies \cite{shapiro2009lectures}. As shown in \cite{rosemberg2021assessing}, time-inconsistency increases power system operational costs, produces higher energy prices and increases emissions. Nevertheless, results from the aforementioned paper indicate that tight convex approximations and relaxations, such as the second-order cone (SOC) relaxation and the Direct-Current with line losses (DCLL) quadratic approximation, can greatly mitigate these negative effects, although still under a significant computational hurdle.

To evaluate the performance of time-inconsistent policies induced by model approximations for long horizons in a tractable manner, a modified version of the SDDP algorithm is used. The modified version performs forward passes using the AC power-flow model and the backward passes, that calculate the future cost functions, using the relaxation - this allows us to fit the future cost functions to the state values that would be visited in the rolling horizon.

\subsection{Additional Results 28-Bus Case} \label{annex: extra_results_bolivia}

This provides additional details and results about the 28-Bus Case.

Policy and optimization details:

\begin{itemize}
    \item All ML models are implemented in Julia using \textit{Flux.jl} \cite{Flux.jl-2018}.
    \item Weights Optimizer: $Adam(\eta = 0.001, \beta::Tuple = (0.9, 0.999), \epsilon = 1.0e-8)$
    \item Latent space size: $64$.
    \item Data split (training batch-size | validation number of samples | test number of samples): $(32 | 1000 | 1000)$.
    \item Experiments are carried out on Intel(R) Xeon(R) Gold 6226 CPU @
        2.70GHz machines with NVIDIA Tesla A100 GPUs on the Phoenix cluster
        \cite{PACE}.
    \item Non-Convex solver: \textit{MadNLP.jl} \cite{shin2023accelerating}.
    \item Quadratic solver: \textit{Gurobi} \cite{gurobi}.
    \item Conic solver: \textit{Mosek} \cite{mosek}.
\end{itemize}

For a broader view of the policy simulation under different models of reality, we show the expected stored energy (volume) and thermal generators dispatch over time for each analysed policy in the Bolivian grid under the SOC and DCLL power flow formulation in the bellow figures. 

\begin{figure}[!ht]
\label{fig:comp-dcll}
    \begin{minipage}[t]{0.5\textwidth}
        \centering
        \includegraphics[width=\linewidth]{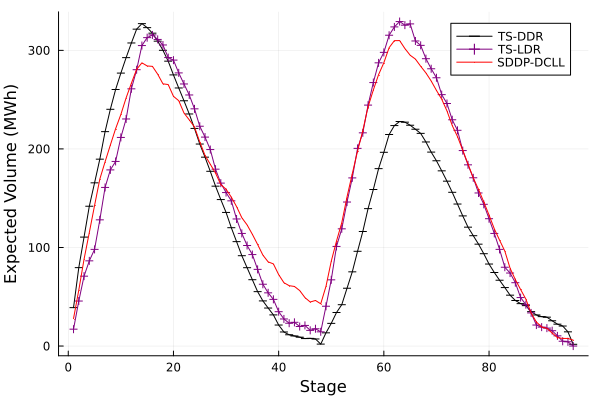}
    \end{minipage}
    \hfill
    \begin{minipage}[t]{0.5\textwidth}
        \centering
        \includegraphics[width=\linewidth]{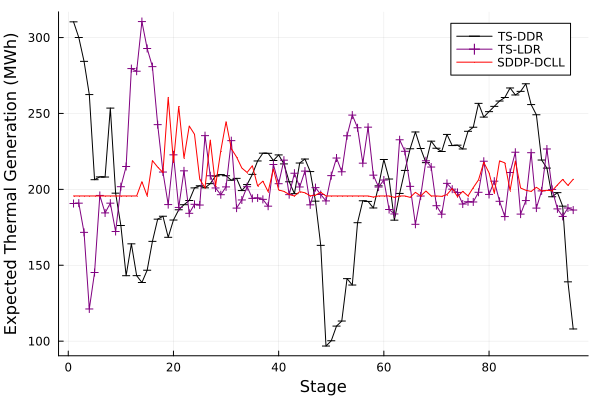}
    \end{minipage}
        \caption{Expected stored energy and thermal dispatch over time for the DCLL formulation.}
\end{figure}

\begin{figure}[!ht]
\label{fig:comp-soc}
    \begin{minipage}[t]{0.5\textwidth}
        \centering
        \includegraphics[width=\linewidth]{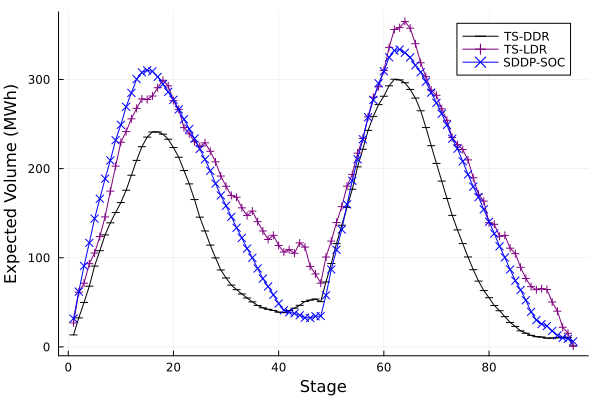}
    \end{minipage}
    \hfill
    \begin{minipage}[t]{0.5\textwidth}
        \centering
        \includegraphics[width=\linewidth]{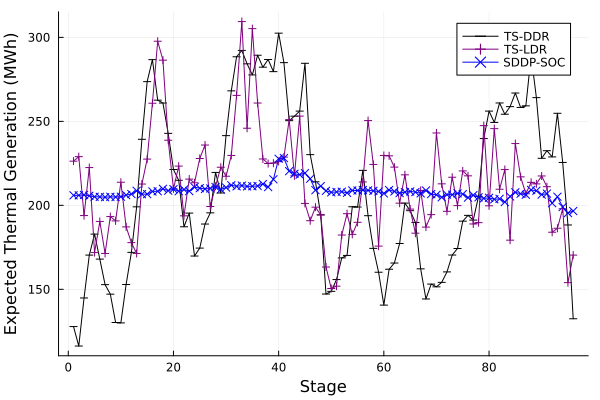}
    \end{minipage}
        \caption{Expected stored energy and thermal dispatch over time for the SOC formulation.}
\end{figure}

\newpage

\subsection{Results on a Small Test Case} \label{annex: results_case3}

This provides details and results for an additional small test case.

Policy and optimization details:

\begin{itemize}
    \item All ML models are implemented in Julia using \textit{Flux.jl} \cite{Flux.jl-2018}.
    \item Weights Optimizer: $Adam(\eta = 0.001, \beta::Tuple = (0.9, 0.999), \epsilon = 1.0e-8)$
    \item Latent space size: $16$.
    \item Data split (training batch-size | validation number of samples | test number of samples): $(32 | 1000 | 1000)$.
    \item Experiments are carried out on Intel(R) Xeon(R) Gold 6226 CPU @
        2.70GHz machines with NVIDIA Tesla A100 GPUs on the Phoenix cluster
        \cite{PACE}.
    \item Non-Convex solver: \textit{MadNLP.jl} \cite{shin2023accelerating}.
    \item Quadratic solver: \textit{Gurobi} \cite{gurobi}.
    \item Conic solver: \textit{Mosek} \cite{mosek}.
\end{itemize}

This case study utilizes a three-bus system to demonstrate the impacts of underlying policies. The system forms a loop by connecting all three buses, making it an illustrative example to observe the influence of Kirchhoff's Voltage Law (KVL) constraints on the quality of the approximations studied. The setup includes a hydroelectric unit at bus 1, a thermo-electric unit at bus 2, and the most costly thermo-electric unit along with the demand at bus 3. The planning horizon spans 48 periods, with each stage representing 730 hours (equivalent to one month). The case study employs three scenarios per stage (low, medium, and high), which sums to $3^{48}$ possible scenarios for the entire problem.

\begin{table*}[!ht]
\centering
\caption{Comparison of SDDP and ML Decision Rule for Case3 with DCLL Implementation.}
\begin{tabular}{ccccccc}
\toprule
\textbf{Model} & \textbf{Plan} & \textbf{Imp Cost (USD)} & \textbf{GAP (\%)} & \textbf{Training (Min)} & \textbf{Execution (Min)} \\
\midrule
SDDP & DCLL & $47703$ & - & - & 3 \\
TS-LDR & DCLL & $48937$ & $2.59$ & 5 & 0.00047 \\
TS-DDR & DCLL & $49449$ & $3.66$ & 5 & 0.00047 \\
\bottomrule
\end{tabular}
\label{tab:case3_dcll}
\end{table*}

\begin{table*}[!ht]
\centering
\caption{Comparison of SDDP and ML Decision Rule for Case3 with SOC Implementation.}
\begin{tabular}{ccccccc}
\toprule
\textbf{Model} & \textbf{Plan} & \textbf{Imp Cost (USD)} & \textbf{GAP (\%)} & \textbf{Training (Min)} & \textbf{Execution (Min)} \\
\midrule
SDDP & SOC & $49178$ & - & - & 17 \\
TS-DDR & SOC & $50819$ & $3.33$ & 30 & 0.0046 \\
TS-LDR & SOC & $50926$ & $3.56$ & 90 & 0.0046 \\
\bottomrule
\end{tabular}
\label{tab:case3_soc}
\end{table*}

\begin{table*}[!t]
\centering
\caption{Comparison of SDDP and ML Decision Rule for Case3 with AC Implementation.}
\begin{tabular}{ccccccc}
\toprule
\textbf{Model} & \textbf{Plan} & \textbf{Imp Cost (USD)} & \textbf{GAP (\%)} & \textbf{Training (Min)} & \textbf{Execution (Min)} \\
\midrule
SDDP & DCLL & $52330$ & - & - & 5 \\
SDDP & SOC & $53705$ & $2.62$ & - & 5 \\
TS-DDR & AC & $53829$ & $2.86$ & 40 & 0.014 \\
TS-LDR & AC & $60493$ & $15.59$ & 150 & 0.014 \\
\bottomrule
\end{tabular}
\label{tab:case3_ac}
\end{table*}

\subsection{RL Algorithms} \label{annex: RL}

The model-free RL algorithms explored are diverse in terms of their underlying approaches, which include both value-based and policy-based methods, as well as hybrid actor-critic approaches:

\begin{itemize}
    \item \textbf{REINFORCE}: A foundational policy-gradient algorithm that directly optimizes the policy by maximizing expected returns through Monte Carlo estimates. REINFORCE updates the policy based on complete trajectories, providing unbiased estimates of the gradient, but can suffer from high variance in environments with complex, high-dimensional action spaces like AC-OPF.


    \item \textbf{Proximal Policy Optimization (PPO)}: A popular actor-critic method that restricts policy updates within a “trust region” by clipping the policy gradient, which stabilizes training. PPO’s robustness and stability make it particularly suited for the high-dimensional feasible region in AC-OPF, where actions need to satisfy strict constraints.



    \item \textbf{Deep Deterministic Policy Gradient (DDPG)}: An off-policy actor-critic algorithm designed for continuous action spaces. DDPG uses a deterministic policy and leverages experience replay and target networks for stability. It is particularly suitable for continuous control problems like AC-OPF but may require significant tuning to handle the complex constraints.

    \item \textbf{Twin Delayed Deep Deterministic Policy Gradient (TD3)}: A variant of DDPG that reduces overestimation bias by using a pair of Q-networks and delaying policy updates. TD3 enhances stability in continuous control tasks and is expected to provide better convergence in the high-dimensional, constrained AC-OPF setting.

    \item \textbf{Soft Actor-Critic (SAC)}: An off-policy entropy-regularized actor-critic method that balances exploration and exploitation by optimizing for maximum entropy in addition to reward. SAC is robust in environments with complex dynamics, making it a strong candidate for navigating the feasible region defined by AC-OPF’s constraints.
\end{itemize}


\end{document}